# A Novel ECOC Algorithm with Centroid Distance Based Soft Coding Scheme

Kaijie Feng; Kunhong Liu(✉); Beizhan Wang

Software School of Xiamen University, Xiamen, China

fengkaijie1995@foxmail.com; lkhqz@xmu.edu.cn; wangbz@xmu.edu.cn;

*Abstract*—**In ECOC framework, the ternary coding strategy is widely deployed in coding process. It relabels classes with {-1, 0, 1}, where -1/1 means to assign the corresponding classes to the negative/positive group, and label 0 leads to ignore the corresponding classes in the training process. However, the application of hard labels may lose some information about the tendency of class distributions. Instead, we propose a Centroid distance-based Soft coding scheme to indicate such tendency, named as CSECOC. In our algorithm, Sequential Forward Floating Selection (SFFS) is applied to search an optimal class assignment by minimizing the ratio of intra-group and inter-group distance. In this way, a hard coding matrix is generated initially. Then we propose a measure, named as coverage, to describe the probability of a sample in a class falling to a correct group. The coverage of a class a group replace the corresponding hard element, so as to form a soft coding matrix. Compared with the hard ones, such soft elements can reflect the tendency of a class belonging to positive or negative group. Instead of classifiers, regressors are used as base learners in this algorithm. To the best of our knowledge, it is the first time that soft coding scheme has been proposed. The results on five UCI datasets show that compared with some state-of-art ECOC algorithms, our algorithm can produce comparable or better classification accuracy with small scale ensembles.**

## I. INTRODUCTION

Nowadays, the multi-class classification problem has been a significant issue in the field of the pattern recognition and machine learning[1]. Usually, there are two popular solutions: the first is the application of a classifier capable of dealing with multi-class problem directly; another is to decompose a multi-class problem into multiple binary-class problems. As it is found that a single classifier can't guarantee high performances for some hard problems, the latter is a more feasible way.

Currently, there are some widely used approaches, such as One-versus-one (OVO), one-versus-all (OVA) and error correcting output codes (ECOC). In detail, OVO combines two classes to form a binary class problem and ignores all other classes in turn[2]; OVA considers one class as the positive group and all other classes as the negative group[3, 4]. Both methods deploy the majority voting scheme to decide the final labels. However, it was proved that ECOC can reduce bias and variance errors produced by the binary classifiers more effectively compared with OVA and OVO.

ECOC is a more general framework, allowing classes to be relabeled according to a coding strategy [5]. With the widely deployed ternary coding strategy, an ECOC algorithm mainly includes two steps: encoding and decoding. In the encoding phase, a $N_C*N$ coding matrix is created with elements taking value {-1, 0, 1}, where $N_C$ represents the number of class. For the coding matrix, each column represents a class partition scheme, matching a base learner. The classes labeled 1/-1 are assigned to positive/negative group, and those labeled 0 are ignored during training process. In the decoding phase, $N$ base learners produce $N$ labels for an unknown sample. And the vector consisting of predicted labels is compared with each row in the coding matrix. The class with the least loss, such as Hamming distance and Euclidean distance, is set as the final result[6].

There are mainly two categories of ECOC algorithms: problem-independent and problem-dependent. Dense Random ECOC (DRECOC) and Sparse Random ECOC (SRECOC) algorithms are problem-independent designs, as they do not take data distribution into account when generating the coding matrix. On the contrary, problem-dependent algorithms, such as DECOC[7], Forest-ECOC[8] and ECOC-ONE[9], take the intrinsic characteristics of data into the consideration. In most cases, it is found that problem-dependent algorithms are superior to the former.

However, the ternary coding scheme neglects the probability of each class belonging to positive group or negative group[10, 11]. And up to now, there is no solution for it in both problem dependent and independent designs. In this paper, we propose a Centroid distance based Soft coding ECOC algorithm, named as CSECOC. It aims to utilize the probability to improve the performance. In this algorithm, classes are allocated to one group at first. Then based on the distances among the centroids of classes and groups, some classes are assigned to another group by maximizing the ratio of inter-group distance to intra-group distance with a Sequential Forward Floating Selection (SFFS) algorithm. After the class assignment scheme is settled, a measure is defined to describe data distribution within each group, named as coverage. It is calculated by the proportion of samples whose nearest centroids matching the correct group. Then coverage is deployed as the membership of a class to its group, taking values in the range of [-1, 1]. By setting the elements in coding matrix with the corresponding coverages, the coding matrix consists of real values depicting the probability of a class belonging to a group. Our ECOC algorithm aims to fit data better, so as to improve the generalization ability.

This paper is organized as follow. Section 2 presents the detail of CSECOC. Section 3 presents the experiment results along with some discussions, and Section 4 concludes this paper.

## II. THE FRAMEWORK OF CSECOC

This section gives the detailed steps of CSECOC in 错误!未找到引用源。. SFFS algorithm [12] is employed to search the best binary partitions by maximizing the ratio of the intra-group distance over inner-group distance as criteria.

Assume for training data set, there are m samples $X = \{x_1 \cdots x_m\}$ with $N_C$ classes, and $C = \{C_1, C_2, \cdots, C_{N_c}\}$ represents the class label set. Let $L$ be the total number of features, and $x_i^l$ represent the $l$-th feature of $x_i$, where $l \in [1, \ldots L]$. Assume the $i$-th class contains $N_{C_i}$ samples. The original class set G would be divided into a binary partition $G_1^k/G_2^k$, including $T_1^k/T_2^k$ classes respectively, where k denotes the times of iteration. Let $center_1^k/center_2^k$ represents the centroid of $G_1^k/G_2^k$, and $center_{C_i}$ stands for the centroid of class $C_i$. In the calculation, these centroids are the mean values of overall samples in a group or class.

$$d(x_i, x_j) = \sqrt{\sum_{l=1}^{L}(x_i^l - x_j^l)^2} \tag{1}$$

$$S(G_i^k) = \begin{cases} \frac{2}{T_i^{k^2} - T_i^k} \sum_{p \neq q; C_p, C_q \in G_i} d(center_{C_p}, center_{C_q}), T_i^k \neq 1 \text{ and } T_i^k \neq N_c - 1 \\ 0, otherwise \end{cases} \tag{2}$$

$$E(G_1^k, G_2^k) = \begin{cases} \frac{d(center_1^k, center_2^k)}{S(G_1^k) + S(G_2^k)}, T_1^k \neq 1 \text{ and } T_2^k \neq N_c - 1 \\ 0, otherwise \end{cases} \tag{3}$$

$$I(x_i) = \begin{cases} 1, & d(x_i, center_1^k) \leq d(x_i, center_2^k) \\ 0, & d(x_i, center_1^k) > d(x_i, center_2^k) \end{cases} \tag{4}$$

$$coverage_{r,l} = \begin{cases} \frac{\sum_{x_j \in C_r} I(x_j)}{N_{C_i}}, if\ C_r \in G_1 \\ -\frac{\sum_{x_j \in C_r} 1 - I(x_j)}{N_{C_i}}, if\ C_r \in G_2 \\ 0, otherwise \end{cases} \tag{5}$$

The inner-group distance and intra-group distance are proposed here to describe two relationships: (1) the relationship among classes in one group; (2) the relationship of classes between two groups. It is intuitive that a large intra-group distance suggests a wide margin between two groups, facilitating the learning task of each base learner. A large inner-group distance paves the way for the next iteration. The inner-group distance is calculated based on the centroids of all classes in a group, while intra-group refers to the distance between centroids of both groups. $S(G_i^k)$ in formula (2) evaluates the inner-group distance for group $G_i^k$. There could be $T_i^k(T_i^k - 1)/2$ possible ways for $T_i^k$ classes in $G_i^k$. And $E(G_1^k, G_2^k)$ in formula (3) calculates the ratio of the inner-group distance and intra-group distance. As a larger $E(G_1^k, G_2^k)$ offers a large margin between groups, it is used as the optimal objective for the searching algorithm.

Here, a measure *coverage* is defined to describe the probability of a sample falling into a true group. Coverage is the proportion of samples in a class whose nearest centroid is the correct group. And formula (5) calculates the coverage of class $r$. It is obvious that if a large proportion of samples in a class can be assigned to a group the class truly belonging to, then the reliability of this assignment scheme is higher. On the contrary, a small coverage reveals an unreasonable class assignment to a group. $I(x_i)$ is used to record to which group sample $x_i$ belongs. By setting the coverage of the $r$-th class for the $l$-th classifier as an element $M_{r,l}$, the final coding matrix contains the membership of every class. The details of CSECOC is shown in 错误!未找到引用源。.

---

**INPUT:** $\{C_1, C_2, \cdots, C_{N_c}\}$
**OUTPUT:** Coding matrix $M$
**Initialization:**
$G_1^0 = \emptyset; G_2^0 = G; G = \{C_1, C_2, \cdots, C_{N_c}\}; k = 0; col = 1.$

**Step 1. search the *k*-th binary partition $G_1^k/G_2^k$ of G:**
    **1.1. Inclusion:**
    $C^+ = argmax_{C_i \in G_2^k} E(G_1^k + C_i, G_2^k - C_i)$
    $G_1^{k+1} = G_1^k + C^+$
    $G_2^{k+1} = G_2^k - C^+$
    **1.2. Conditional exclusion:**
    $C^- = argmax_{C_i \in G_1^k} E(G_1^k - C_i, G_2^k + C_i)$
    IF $E(G_1^k - C^-, G_2^k + C^-) > E(G_1^k, G_2^k)$ THEN
        $G_1^{k+1} = G_1^k - C^-$
        $G_2^{k+1} = G_2^k + C^-$
        go to **Step1.2**
    ELSE IF $E(G_1^k, G_2^k) \approx E(G_1^{k-1}, G_2^{k-1})$ THEN
        $k = k + 1$
        go to **Step2**
    ELSE
        go to **Step1.1**
**Step 2. Calculate the *col*-th column in coding matrix $M$:**
    FOR $i \in \{1, 2, \cdots, N_c\}$
        $M_{i, col} = coverage_{i, col}$
    $col = col + 1$
**Step 3. coding the $G_1^k$ and $G_2^k$**
    IF $|G_i^k| > 1\ i \in [1,2]$ THEN
        $G = G_i^k$

Fig 1. the detail of CSECOC algorithm

In 错误!未找到引用源。, step 1 employs SFFS to find the best binary partition $G_1^0/G_2^0$ for the original group $G$. The initialized partition is $G_1^0$ and $G_2^0$, where $G_1^0$ is empty and $G_2^0$ contains all classes in the training set. At step1.1, the algorithm tries to remove a class $C^+$ from $G_2^0$ and add it to $G_1^0$ to maximize the evaluation function $E(G_1^0, G_2^0)$. At step1.2, the worst class $C^-$ is moved from $G_1^0$ to $G_2^0$ to keep criterion increasing. If the new partition is better than the original partition, then new partitions would be accepted. Otherwise our algorithm goes back to step1.1. Step 1 is terminated when $E(G_1^0, G_2^0)$ cannot be increased anymore.

Step 2 aims to fill the coding matrix by means of estimating the probability of each class belonging to positive or negative group. For each class, Euclidean distance measurement is used to calculate the distance to two group centroids for each sample. After this step, both $G_1^0$ and $G_2^0$ are checked. If one of them contains more than one class, the division process will continue. The algorithm stops only when each group consists of one class. In general, for a $N_C$ class problem, the algorithm repeats this splitting process $N_C$-1 times, and each iteration contributes a new column to the coding matrix. So our algorithm produce a coding matrix with $N_C$-1 columns.

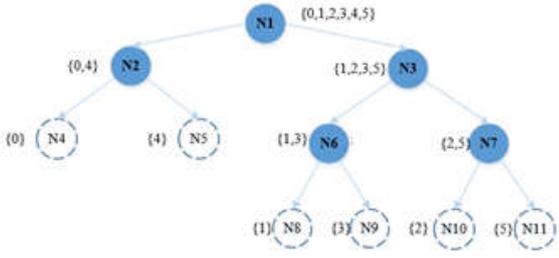

Fig 2. Process of creating coding matrix

Fig 3. Coding Matrix based on 错误!未找到引用源。

The class partition process can be mapped as a binary tree, and the creation of a coding matrix is illustrated as 错误!未找到引用源。. Here the left child node represents positive group and the right child node represents negative group. The root node N1 represents six classes, {0,1,2,3,4,5}. Assume that $E(G_1^0, G_2^0)$ is maximized by assigning {0, 4} to the positive group and {1, 2, 3, 5} to the negative group, then the coverage of each class is calculated. The classes assigned to positive group will obtain positive coverage values, and the classes belonging to negative group can only receive negative values. This class decomposition process repeats, and produces a coding matrix with five columns, as shown in 错误!未找到引用源。.

In Fig. 3, $\{C_0, C_1, \cdots, C_5\}$ represent different classes and $\{H_0, H_1, \cdots, H_4\}$ represent different base learners for each column. As each element show the membership of corresponding class belonging to a group, from the first column, it is found that the samples of $C_1, C_2, C_3$ belong to the negative group at 100%. Only 91% samples in $C_5$ belong to the negative group, and the remaining 9% samples close to the positive group instead. In another hand, only 73% samples in $C_4$ close to the centroid of positive group. So such soft coding scheme provides us more information about data distributions in different classes.

III. EXPERIMENTS AND DISCUSSIONS

In the experiment, five datasets from UCI repository are deployed to validate our approach, and the details are listed in TABLE I.

We compare our approach with OVO, OVA, DRECOC, SRECOC and DECOC. Because of different working principle, different ECOC algorithms produce different coding matrices with various sizes, as shown in TABLE II. DECOC uses the formula(3) as evaluation function to create hierarchical structure. In our algorithm, the targets are real values instead of class labels. So two regressors are deployed as base learners, SVR with RBF kernel [13]and KNN (K=5) based regressor. As other ECOC algorithms require classifiers as base learners, SVM and KNN are applied instead. These base learners are picked from Scikit-learn toolbox with default settings[14].

In experiments, stratified 10-fold cross validation is applied ten times with random splitting. So the mean accuracies and standard deviation results are listed in Table II and Table III, and the highest accuracies are marked in bold fond. To simplify our discussions, all features are used in the whole process.

TABLE I. Description of the data sets used in experiments

| #Index | #Name | #Samples | #Features | #Classes |
|---|---|---|---|---|
| A | Dermatology | 358 | 34 | 6 |
| B | Wine | 178 | 13 | 3 |
| C | Iris | 150 | 4 | 3 |
| D | Thyroid | 215 | 5 | 3 |
| E | Vehicle | 846 | 18 | 4 |

TABLE II. Ensemble sizes of different algorithms

| OVO | DRECOC | SRECOC | OVA, DECOC, CL-ECOC |
|---|---|---|---|
| $(N_c - 1) \times N_c$ | $10 \log N_c$ | $15 \log N_c$ | $N_c - 1$ |

Fscore and accuracy are two widely used measures for the evaluation of different algorithms performances. The original Fscore and accuracy are designed for binary problems. When applied in a multiclass problem, the average Fscore and accuracy among classes are used. That is, for the i-th binary problem, the i-th class is regard as the positive class, and others are labelled as the negative class, so positive rate ($P_i$), negative rate ($N_i$), true positive ($TP_i$), true negative ($TN_i$), false positive ($FP_i$) and false negative ($FN_i$) are calculated as those in a binary problem. The final score is the average of all binary problems, as shown by formulas (6-9), and $\beta$ is set to 1 to get balanced results.

$$\text{Accuracy} = \text{avg}(\sum_{i=1}^{R} \frac{TP_i + TN_i}{P_i + N_i}) \quad (6)$$

$$\text{Precision} = \text{avg}(\sum_{i=1}^{R} \frac{TP_i}{TP_i + FP_i}) \quad (7)$$

$$\text{Recall} = \text{avg}(\sum_{i=1}^{R} \frac{TP_i}{P_i}) \quad (8)$$

$$\text{Fscore} = \text{avg}(\sum_{i=1}^{R} \frac{(\beta^2+1) * Presicion_i * Recall_i}{\beta^2 * Presicion_i + Recall_i}) \quad (9)$$

From TABLE III and TABLE IV, it is found in general, problem-dependent ECOC algorithms can beat problem-independent ones in most cases. That is, DECOC and CSECOC algorithm can achieve high accuracies and Fscores in most cases with the smallest ensemble size. As DRECOC and SRECOC are based on random coding algorithms, their performances are not so stable, and SRECOC never wins in experiments.

With SVM/KNN as base learner, our algorithm wins two/three out of five cases based on average accuracies, and achieves the highest average mean results. Even though OVO employs much more base learners compared with our algorithm, its performance slightly worse. The same conclusions can be drawn from the results of Fscores. So the advantage of our algorithm is obvious. And it should be noted that because our algorithm requires less learners, it can be trained and tested faster.

TABLE III. Results of different methods using SVM

| Measures | datasets | OVA | OVO | DRECOC | SRECOC | DECOC | CSECOC |
|---|---|---|---|---|---|---|---|
| Accuracy | A | 95.2±1.70 | 97.2±1.49 | 96.3±1.37 | 89.9±7.04 | **97.5±0.93** | 97.4±0.81 |
| | B | 96.9±1.45 | 97.4±1.89 | **98.5±0.74** | 98.1±1.43 | 98.1±1.17 | 97.8±1.39 |
| | C | 95.8±2.32 | 95.5±2.81 | 96.0±2.40 | 95.8±2.71 | **96.8±1.78** | 96.2±2.23 |
| | D | 95.2±2.96 | 95.5±2.62 | 94.5±2.50 | 94.2±2.98 | 95.3±1.95 | **96.9±1.69** |
| | E | 61.6±2.26 | 76.0±1.51 | 72.8±4.21 | 75.9±2.80 | 75.1±2.53 | **78.7±2.59** |
| | mean | 88.9±2.14 | 92.3±2.06 | 91.6±2.24 | 90.8±3.39 | 92.6±1.67 | **93.4±1.74** |
| Fscore | A | 95.0±1.83 | 97.2±1.41 | 96.3±1.43 | 88.2±8.30 | **98.2±1.42** | 97.4±0.81 |
| | B | 96.8±1.50 | 97.4±1.90 | **98.5±0.71** | 98.1±1.42 | 98.3±1.20 | 97.8±1.42 |
| | C | 95.7±2.35 | 95.6±2.80 | 96.0±2.33 | 95.8±27.11 | **97.1±2.61** | 96.3±2.23 |
| | D | 95.0±3.27 | 95.3±2.82 | 94.2±2.71 | 93.8±3.23 | 95.0±2.35 | **96.8±1.88** |
| | E | 58.5±2.30 | 74.9±1.93 | 70.8±5.32 | 74.8±3.02 | 75.3±3.26 | **78.4±2.81** |
| | mean | 88.2±2.69 | 92.1±2.17 | 91.2±2.50 | 90.1±8.62 | 92.8±2.17 | **93.3±1.83** |

TABLE IV. Results of different methods using KNN

| Measures | datasets | OVA | OVO | DRECOC | SRECOC | DECOC | CSECOC |
|---|---|---|---|---|---|---|---|
| Accuracy | A | 94.4±1.71 | 94.9±1.86 | **95.6±1.16** | 87.7±10.03 | 94.4±1.71 | 93.5±3.70 |
|  | B | 97.2±1.71 | 97.2±1.71 | 95.7±1.67 | 96.1±1.93 | 97.2±1.71 | **98.1±0.83** |
|  | C | 94.7±2.47 | 94.7±2.47 | **95.3±2.71** | 94.9±2.00 | 94.6±3.01 | 94.7±2.47 |
|  | D | 92.2±3.11 | 92.2±3.11 | 94.0±2.96 | 94.0±4.32 | 92.0±3.14 | **94.2±3.88** |
|  | E | 68.4±2.05 | 71.5±2.45 | 70.6±1.39 | 70.4±3.36 | 71.1±2.31 | **71.5±2.72** |
|  | mean | 89.4±2.21 | 90.1±2.32 | 90.2±2.30 | 88.6±5.12 | 89.9±2.38 | **90.4±2.72** |
| Fscore | A | 94.5±1.71 | 95.0±1.89 | **95.7±1.11** | 86.5±11.0 | 95.1±2.31 | 93.1±5.01 |
|  | B | 97.2±1.72 | 97.2±1.74 | 95.7±1.72 | 96.1±2.01 | 97.5±2.22 | **98.2±0.81** |
|  | C | 94.7±2.53 | 94.7±2.51 | **95.4±2.71** | 94.9±2.03 | 95.3±3.13 | 94.7±2.53 |
|  | D | 91.6±3.49 | 91.6±3.42 | 93.7±3.28 | 93.6±4.76 | 91.2±3.62 | **94.0±4.12** |
|  | E | 66.7±2.53 | 70.1±2.81 | 69.9±1.89 | 69.4±3.74 | 70.2±3.71 | **70.3±3.11** |
|  | mean | 88.9±2.40 | 89.7±2.47 | 90.1±2.14 | 88.1±4.71 | 90.0±2.99 | **90.1±3.11** |

## IV. CONCLUSION

In this paper, we introduced a new ECOC based on the evaluation of centroid loss. It split classes into two partitions by minimizing the ratio of intra-group and inter-group. After the class assignment is settled, the algorithm evaluates the probability of each class belonging to positive group or negative group, which is deployed as elements in the coding matrix. In this way, our algorithm produce soft coding matrix, and each element takes value within $\{-1,1\}$. Comparing with other methods, our algorithm can achieve the best average accuracy and Fscores, and obtain the highest accuracy and Fscores with SVM and KNN in most cases.

It is obvious that there are still some more topics concerning with this algorithm, such as some new manners to define the soft codes, in ECOC algorithm. It is our future research direction.

ACKNOWLEDGMENT

This work is supported by National Key Technology Research and Development Program of the Ministry of Science and Technology of China (2015BAH55F05); Natural Science Foundation of Fujian Province (No. 2016J01320 and 2015J05129), and National Natural Science Foundation of China (Grant No.61502402 and 61772023).